\documentclass{elsarticle}

\usepackage[utf8]{inputenc}
\usepackage{subfig}
\usepackage{url}
\usepackage{my_symbols}
\usepackage[show]{chato-notes}
\usepackage{color}
\usepackage[dvipsnames]{xcolor}
\usepackage{tikz} 
\usepackage{pgfplots, pgfplotstable}
\usepackage{blindtext}
\usepackage{tabularx}
\usepackage{chngpage}

\colorlet{maxcolor}{OliveGreen}

\usetikzlibrary{calc,trees,positioning,arrows,chains,shapes.geometric,%
    decorations.pathreplacing,decorations.pathmorphing,shapes,%
    matrix,shapes.symbols,automata}

\makeatletter
\pgfplotsset{
    /tikz/max node/.style={
    },
    /tikz/min node/.style={
        anchor=south west,
        name=minimum
    },
    mark min/.style={
        point meta rel=per plot,
        visualization depends on={x \as \xvalue},
        scatter/@pre marker code/.code={%
            \ifx\pgfplotspointmeta\pgfplots@metamin
                \def\markopts{}%
                \coordinate (minimum);
                \node [min node] {
                    (\pgfmathprintnumber[fixed]{\xvalue},%
                    \pgfmathprintnumber[fixed]{\pgfplotspointmeta})
                };
            \else
                \def\markopts{mark=none}
            \fi
            \expandafter\scope\expandafter[\markopts,every node near coord/.style=green]
        },%
        scatter/@post marker code/.code={%
            \endscope
        },
        scatter,
    },
    mark max/.style={
        point meta rel=per plot,
        visualization depends on={y \as \yvalue},
        scatter/@pre marker code/.code={%
        \ifx\pgfplotspointmeta\pgfplots@metamax
            \def\markopts{}%
            \coordinate (maximum);
            \node [max node] {
                (\pgfmathprintnumber[fixed]{\yvalue})
            };
        \else
            \def\markopts{mark=none}
        \fi
            \expandafter\scope\expandafter[\markopts]
        },%
        scatter/@post marker code/.code={%
            \endscope
        },
        scatter
    }
}
\makeatother

\begin{document}

\begin{frontmatter}

\title{Data Stream Classification using Random Feature Functions and Novel Method Combinations}

\author[BSC]{Diego Marrón}
\ead{dmarron@ac.upc.edu}

\author[AALTO]{Jesse Read}
\ead{jesse.read@aalto.fi}

\author[TELECOM]{Albert Bifet}
\ead{albert.bifet@telecom-paristech.fr}

\author[BSC]{Nacho Navarro}
\ead{nacho@ac.upc.edu}

\address[BSC]{Department of Computer Architecture, Universitat Politecnica de Catalunya and with the Department of Computer Science, Barcelona Supercomputing Center, Spain}
\address[AALTO]{Aalto University and HIIT,Finland}
\address[TELECOM]{Télécom ParisTech, Paris, France}


%
%

\begin{abstract}

Big Data streams are being generated in a faster, bigger, and more commonplace. 
In this scenario, Hoeffding Trees are an established method for classification. Several 
extensions exist, including high-performing ensemble setups such as online and leveraging
bagging. Also, $k$-nearest neighbors is a popular choice, with most extensions dealing with
the inherent performance limitations over a potentially-infinite stream.

At the same time, gradient descent methods are becoming increasingly popular, owing in part
to the successes of deep learning. Although deep neural networks
can learn incrementally, they have so far proved too sensitive to hyper-parameter options and 
initial conditions to be considered an effective `off-the-shelf' data-streams solution.

In this work, we look at combinations of Hoeffding-trees, nearest neighbour, and gradient descent
methods with a streaming preprocessing approach in the form of a random feature functions filter 
for additional predictive power. 

We further extend the investigation to implementing methods on GPUs, which we test on some large 
real-world datasets, and show the benefits of using GPUs for data-stream learning due to their high
scalability.

Our empirical evaluation yields positive results for the novel approaches that we experiment with, highlighting  important issues, and shed light on promising future directions in approaches to data-stream classification.

\end{abstract}

%
%

\end{frontmatter}

%
%
\section{Introduction}

There is a trend towards working with big and dynamic data sources. This tendency is clear both in real world applications and the academic literature. Many modern data sources are not only dynamic but often generated at high speed and must be classified in real time. Such contexts can be found in sensor applications (e.g., tracking and activity monitoring), demand prediction (e.g., of electricity), manufacturing processes, robotics, email, news feeds, and social networks. Real-time analysis of data streams is becoming a key area of data mining research as the number of applications in this area grows. 

The requirements for a classifier in a data stream are to
\begin{itemize}
	\item Be able to make a classification at any time
	\item Deal with a potentially infinite number of examples
	\item Access each example in the stream just once
\end{itemize}

These requirements can in fact be met by variety of learning schemes, including even batch learners (e.g., \cite{QuZZQ09}), where batches are constantly gathered over time, and newer models replace older ones as memory fills up. Nevertheless, incremental methods remain strongly preferred in the data streams literature, and particularly the Hoeffding tree (HT) and its variations \cite{DomingosH00, BifetHP10}, $k$-nearest neighbors ($k$NN) \cite{EykeIBL}. Support for these options is given by large-scale empirical comparisons \cite{IDA2012}, where it is also found that methods such as naive Bayes and stochastic gradient descent-based (SGD) are relatively poor performers.

%


%
%

Classification in data streams is a major area of research, in which Hoeffding trees have long been a favoured method. The main contribution of this paper is to show that random feature function can be leveraged by other algorithms to obtain similar or even improved performance over tree-based methods. 
 

With the recent popularity of Deep Learning (DL) methods we also want to test how a random feature in the form of random projection layer performs on Deep Neural Networks (DNNs).


DL aims for a better data representation at multiple layers of abstraction, and for each layer the network needs to be fine-tuned. In classification, a common algorithm to fine-tune the network is the SGD which tries to minimize the error at the output layer using an objective function, such as Mean Squared Error (MSE). A Gradient vector is used to back-propagate the error to previous layers. This gradient nature of the algorithm makes it suitable to be trained incrementally in batches of size one, similar to how incremental training is done. Unfortunately, DNN are very sensitive to hyper-parameters such as learning rate ($\eta$), momentum ($\mu)$, number of number neurons per level, or the number of levels. It is then not straight forward to provide an of-the-shelf method for data streams.

%

Propagation between layers is usually done in the form of matrix-vector or matrix-matrix multiplications, which are computational intensive operation. Often hardware accelerators such as FPGAs or GPUs are used to accelerate the calculations. Despite some efforts, acceleration of HT and $k$NN algorithms for data streams on the GPUs are has some limitations. We talk briefly about this in Section \ref{sec:related}.

In recent years, Extreme Learning Machines \cite{Huang15} (ELMs) have emerged as a popular framework in Machine Learning. ELMs are a type of feed-forward neural networks characterized by a random initialization of their hidden layer, combined with a fast training algorithm. Our random feature method is based on this approach.




We made use of the MOA (Massive Online Analysis) framework~\cite{MOA}, a software environment for implementing algorithms and running experiments for online learning from data streams in Java. It implements a large number of modern methods for classification in streams, including HT, $k$NN, and  SGD-based methods. We make use of MOA's extensive library of methods to form novel combinations with these methods and further employ an extremely rapid preprocessing technique of projecting the input into a new space via random feature functions (similar to ELMs). We then took the methods purely related to Neural Networks (those which proved most promising under random projections) and implemented them using NVIDIA GPUs and CUDA 7.0; comparing performance to the methods in MOA.

This paper is organized as follows: Section \ref{sec:related} introduces related work on tree based approaches, neural networks, and data streams on GPU. We discuss the use of random features  in Sections \ref{sec:treebased} and \ref{sec:randomlayer} for HT/$k$NN methods and neural networks respectively. We first present the evaluation of tree-based methods in Section \ref{sec:exp:tree} and later in Section \ref{sec:exp:dnn} we extend the SGD method in the form of DNNs, using different activation functions. We finally conclude the paper in Section \ref{sec:conclusion}.

%
%
\section{Related Work}
\label{sec:related}

Hoeffding trees~\cite{DomingosH00} are state-of-the-art in classification for data streams and they predict by choosing the majority class at each
leaf. 
However, these trees may be conservative at first and in many situations naive Bayes method outperforms the standard Hoeffding tree initially, although it is eventually overtaken \cite{HolmesKP05}. A proposed hybrid adaptive method (by \cite{HolmesKP05}) is a Hoeffding tree with naive Bayes at the leaves, i.e., returning a naive Bayes prediction at the leaves, if it has been so far more accurate overall than the majority class. Given it's widespread acceptance, this is the default in MOA, and we denote this method in the experimental Section simply as HT. In fact, the naive Bayes classification comes for free, since it can be made with the same statistics that are collected anyway by the tree. 

Other established examples include using principal component analysis (reviewed also in \cite{EoSL}) for this transformation, and also Restricted Boltzmann Machines (RBMs) \cite{DBNbp}. RBMs can be seen as a probabilistic binary version of PCA, for finding higher-level feature representations. They have received widespread popularity in recent years due to their use in successful deep learning approaches. In this case, $\z = \phi(\x) = f(\W^\top \x)$ for some non-linearity $f$: a sigmoid function is typical, but more recently rectified linear units (ReLUs, \cite{ReLU}) have fallen into favour. The weight matrix $\W$ is learned with gradient-based methods \cite{CD}, and the projected output should provide a better feature representation for a neural network or any off-the-shelf method. This approach was applied to data streams already in \cite{SAC2015}, but concluded that the sensitivity to hyper-parameters and initial conditions prevented good `out-of-the-box' deployment in data streams.

Approaches such as the so-called extreme learning machines (ELMs) \cite{ELM} avoid tricky parametrizations by simply using random functions (indeed, ELMs are basically linear learners on top of non-linear data transformations). Despite the hidden layer weights being random , it has been proven that ELMs is still capable of universal approximation of any non-constant piecewise contiuous function \cite{Huang06}.

Also an incremental version of ELMs is proposed in \cite{Huang07}. It starts with an small network, and new neurons are added at each step until an stopping criterion of size or residual error is reached. The difference with our incremental build is that we use one instance at time simulating they arrive in time, and we incrementally train the network. Also our number of neurons is fixed during the training, in other words, we don't add/remove any neuron during the process.

Nowadays, in 2015, it is difficult when talking about DL and DNNs not to mention GPUs. They are a massive parallel architectures providing an outstanding performance for High Performance Computing and a very good performance/watt ratio, as their architecture suits very fine to their needs of DNNs computations. Many tools includes a back-end to offload the computation to the GPU. NVIDIA has its own portal for deep learning on GPUs at https://developer.nvidia.com/deep-learning.

GPUs has not only used to accelerate DL/DNN computations due to its performance, it has been also been used to successfully accelerate HT and ensembles. However, few works are provided in the context of data streams and GPUs. 

The only work we are aware of regarding to HT in the context of online real-time data streams mining is\cite{MarronECAI14}, were the authors present a parallel implementation of HT and Random Forests for binary trees and data streams achieving goods speedups, but with limitations on the size and with high memory consumption. More generic HT implementation of Random Forests is presented in \cite{GrahnLLS11}. In \cite{SchulzWSB15} the authors introduced an open source library, available at github, to predict images labelling using random forests. The library is also tested their on a cell phone with VGA resolution in real-time with good results.

Also, $k$NN has already been successfully ported to GPUs \cite{Garcia08fastk}. That paper presented one of the first implementations of the ``brute force" $k$NN on GPUs, and compared with several CPU-based implementations with speedups up to teo orders of magnitude. $k$NN is also used in business intelligence \cite{6386646} and has also its implementation on the GPU. The same way as with HT, a tool for machine learning (including $k$NN) is described in \cite{Liu:2015:PPM:2694344.2694358}.

%
%
\section{Tree Based Random Feature Functions}
\label{sec:treebased}

Transforming the feature space prior to learning and classification is an established idea in the statistical and machine learning literature \cite{EoSL}, for example with basis (or feature-) functions. Suppose the input instance is $\x$ of length $d$. This vector is transformed to a new space $ \z = \phi(\x) $ via function $\phi$, creating new vector $\z$ of length $h$. Any off-the-shelf model now treats $\z$ as if it were the input. The functions can be either chosen suitably by a domain expert, or simply chosen to achieve a more dimensioned representation of the input. Polynomials and splines are a typical choice.

Regarding HTs with additional algorithms in the leaves (as described in \Sec{sec:related}), this filter can either be placed before the HT, or before the method in the leaves, or both. 

In this paper we adapt this methodology to deal with other classifiers in a similar way, namely $k$NN and an SGD-based method (rather than naive Bayes) at the leaves. We denote these cases HT-$k$NN and HT-SGD, respectively. For example, in HT-SGD, a gradient descent learner is employed in the leaves of each tree.  Similarly to HT, predictions by the $k$NN and an SGD-based method are only used if they are more accurate on average than the majority class.


\subsection{Ensembles in Data Streams}

\def\adwin{{\tt ADWIN }}
Bagging is an ensemble method used to improve the accuracy of classifier methods. Non-streaming bagging~\cite{bagging} builds a set of $M$ base models, training each model with a bootstrap sample of size $N$ created by drawing random samples with replacement from the original training set. Each base model's training set contains each of the original training example $K$ times where $P(K=k)$ follows a binomial distribution. This binomial distribution  for large values of  $N$ tends to a Poisson($\lambda =1$) distribution, where  Poisson($\lambda =1$)$= \exp(-1)/k!$. Using this fact, Oza and Russell~\cite{OzaR01,oza01online} proposed {\em Online Bagging}, an online method that instead of sampling with replacement, gives each example a weight according to Poisson(1). \adwin Bagging~\cite{BifetHPKG09} is an adaptive version of Online Bagging that uses a change detector to decide when to discard under-performing  ensemble models. 

Leveraging Bagging (LB,~\cite{BifetHP10}) improves \adwin Bagging, increasing the weights of this resampling using a larger value $\lambda$ to compute the value of the Poisson distribution. The Poisson distribution is used to model the number of events occurring within a given time interval. It proved very competitive.

\begin{figure}
	\centering
	\includegraphics[scale=0.6]{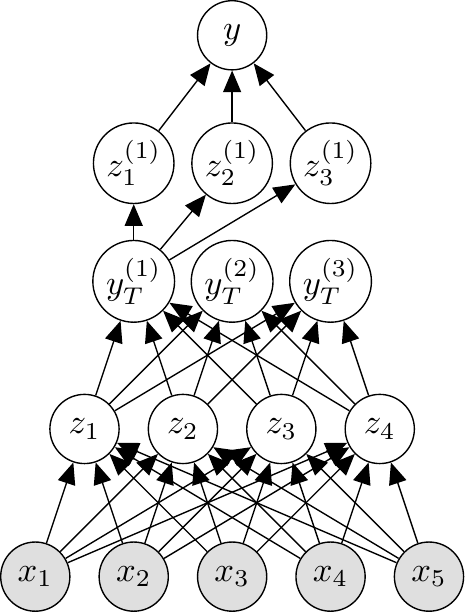}
	\caption{\label{fig:deep}An example setup: Input $\x$ is filtered (i.e., projected) to random layer $\z$ (first layer of connections), which goes to an ensemble of, for example, HTs (second layer), wherein instances are partitioned to the leaves and are again filtered (third layer) and used as training for, say, SGD, producing (in the firth layer of connections) final vote $y$. Note, however, that we only draw the final two layers wrt to the first of the HT models.}
\end{figure}

Again, we can run a filter on the input instances before entering the ensemble of trees, or at the leaves. It is even possible to run the filter again on the \emph{output} of an ensemble (i.e., the votes), before running an additional stacking procedure. This kind of methodology can give way to rather `deep' classifiers.  \Fig{fig:deep} illustrates a possible setup. In this sense of multiple levels we could also call our approach \emph{deep} learning. It is debatable whether decision trees can be called a deep method (their levels involve partitioning an existing feature set rather than because they simple partition a space rather than create higher-level feature space). However, several of the methods we investigate have at least multiple levels of feature transformation, which is behind the power of most deep methods. In the following Section we investigate the empirical performance of several novel combinations based on the methodology described so far.

%
%
\section{Neural Networks with Random Projections for Data Streams}
\label{sec:randomlayer}

\begin{figure}
	\centering
	\includegraphics[scale=0.6]{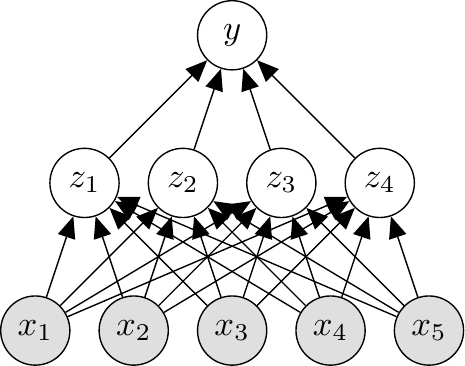}
	\caption{\label{fig:rpl} Random Projection Layer: Input $\x$ is projected to a random layer $\z$ (first layer of connections), which is trained to produce the final vote $y$.}
\end{figure}

Data streams are potentially infinite, and so, they can evolve with time. This means the statistical distribution of the data we are interested on can change. The idea behind the random layer is to improve data localization across the space the trained layer sees. Imagine the instance is a tiny luminous point on the space, with enough random neurons acting as a mirror we hope the trained layer can capture better the data movement. The strategy used by the random projection layer es shown in Figure \ref{fig:rpl}

Except for the fact it is never trained, the random layer is a normal layer and need its activation functions, in this work sigmoid, ReLU, ReLU incremental and a Radial Basis Function are used. 

The sigmoid function used is the standard one, with $\sigma(x) \in [-1,1]$:
$$ 
	\sigma(a_k) = \frac{1}{1+ e^{-a_k}}
$$

where $a_k = \W_k ^\top\x$ is the $k$-th activation function and $\W$ is the weight $d \times h$ matrix ($d$ input attributes, $h$ output features).

\begin{figure}
	\centering
	\includegraphics[scale=0.20,trim=0.0cm 6cm 0.0cm 6cm,clip]{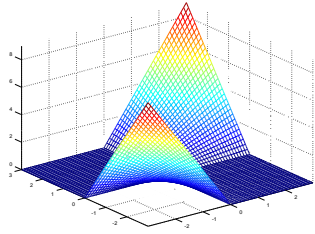}
	\caption{\label{fig:ReLU} Terrain of ReLU basis function on two input attributes $x_1,x_2$ the feature function $z$ is given on the vertical axis.}
\end{figure}

ReLU functions are defined as:
$$
	z_k = f(a_k) =  max(0,a_k)
$$
As stated in Section \ref{sec:related}, ReLUs activation are very efficient as they require only a comparison. In our random projection we expect near 50\% of the neurons to be active for a single instance ( the terrain of a ReLU is exemplified in \Fig{fig:ReLU} ). 

One variation we can do to the standard ReLU is to use the mean value of the attribute as a threshold. The mean value is calculated incrementally as instances arrive. We call this variant ReLU incremental, and is defined as:
$$ 
	f(a_k) = max(\bar{a}_k,a_k)
$$

The last activation function we are using is the Radial Basis Function (RBF):
$$
	\phi(x) = e^{-\frac{(x-c_i)^2}{2\sigma^2}} 
$$

where $x$ is an input instance attribute value and $c_i$ is a random value set at initialization time. Each neuron in the random layer has its own set of $c_i$, the length of both vector $x$ and $c$ are the same. So, we can see the operation $(x-c_i)^2$ as the euclidean distance of the current instance to a randomly positioned center. The $\sigma^2$ is a free parameter. A simplification we can do to this notation is:

$$
\gamma = \frac{1}{2\sigma^2}
$$

In our experiments we try different $\gamma$ values passed at command line. We use the following notation in our experiments:

$$
	\phi(x) = e^{-\gamma{(x-c_i)^2}} 
$$

All matrices and vectors in our model are initialized using random numbers. Matrices are used as normal weight matrix, but the function of the vectors are activation function dependent. Usually initialization is done using random numbers with $\mu=0$ and $\sigma=1$. Assuming our data range $\in [-1,1]$ if we put a Gaussian centered at one of the endpoints, half of its are of influence area if wasted and will never see a point making it harder to fill the whole space and so the discovering of points.

\begin{figure*}
	\centering
	\subfloat[test][Typical Random initialization]{
		\label{fig:rand_a}
		\includegraphics[width=0.25\textwidth]{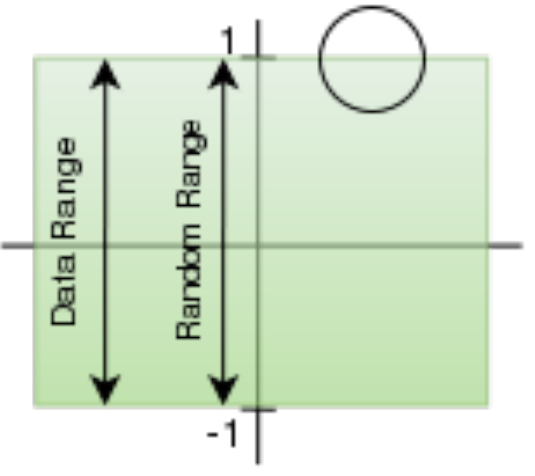}
	}
	\quad
	\subfloat[test][Better use of random range]{
		\label{fig:rand_b}
		\includegraphics[width=0.25\textwidth]{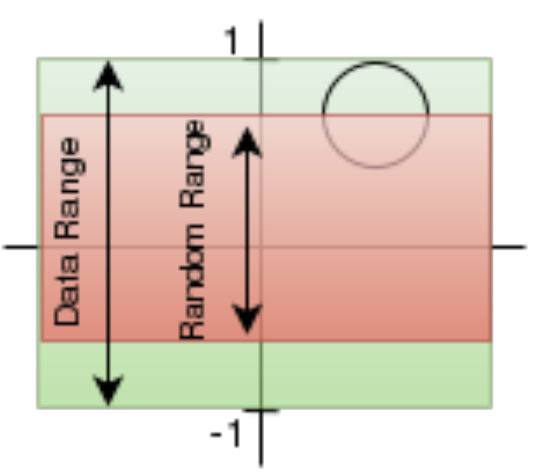}
	}
	\caption{\label{fig:4} Random number initialization strategies}
\end{figure*}

If a smaller range is used, $\sigma \in (0,1)$ (note the open interval), we can improve each neuron's area of influence, as shown in figure \ref{fig:rand_b}. In red the random numbers range is smaller than data range so if we put a Gaussian at the random endpoint can improve its influence are. In this example we used a Gaussian function as an example, but we the idea extends the same for activation functiones. In fact this is what we do in Section \ref{sec:exp:dnn}, specially when talking about the sigmoid neurons as they are always used at the trained layer.

%
%
\section{Random Feature Function Evaluation} 
\label{sec:exp:tree}


Among the methods we investigate (e.g., HT, $k$NN, SGD\footnote{We refer, in this case, to the instantiation with default parameters in MOA, i.e., minimizing hinge loss}), different levels of filters and ensembles and possibly additional classification in the leaves (in the case of HT), there are a multitude of possible combinations. We first investigate the viability of random feature functions and their effect on the different classifiers (comparing these common methods with their `filtered' versions that we denote HT-SGD, $k$NN-F, and SGD-F. This study led us to novel combinations, which we further compare to the benchmark methods and state-of-the-art Leveraging Bagging (LB-HT).

The random feature used in these evaluations are basically ELMs \cite{ELM}. In this Section, we use only ReLU (explained in Section \ref{sec:randomlayer}) as the activation function. In Section \ref{sec:randomlayer} we extend the functions used within the random feature to define a random projectin layer for DNNs. 

Our random feature is based on ELMs, which are defined using Radial Basis Funciontions, but instead in this section we use ReLU as the activation functions. Both functions are defined in detail in Section \ref{sec:randomlayer}

All experiments in this section were carried out using the MOA framework \cite{MOA} with prequential evaluation: each individual example is used to test the model before it is used for training, and from this the accuracy can be incrementally updated. We used an 8-core (3.20GHz each) desktop machine allowing up to 1 gigabyte of RAM per run (all methods were able to finish).

\Tab{tab:datasets} lists the data sources used. A thorough description of most of the datasets is given in \cite{IDA2012}. Of the others, LOC1 and LOC2 are datasets dealing with classifying the location of an object in a grid of $5 \times 5$ and $50 \times 50$ pixels respectively, described in \cite{DCC}. SUSY \cite{SUSY} has features that are kinematic properties measured by particle detectors in an accelerator. The binary class distinguishes between a signal process which produces supersymmetric particles and a background process which does not. It is one of the largest datasets in the UCI repository that we could find.

\begin{figure}
  \begin{center}
    \subfloat[test][Accuracy]{
	\label{fig:3a}
	\includegraphics{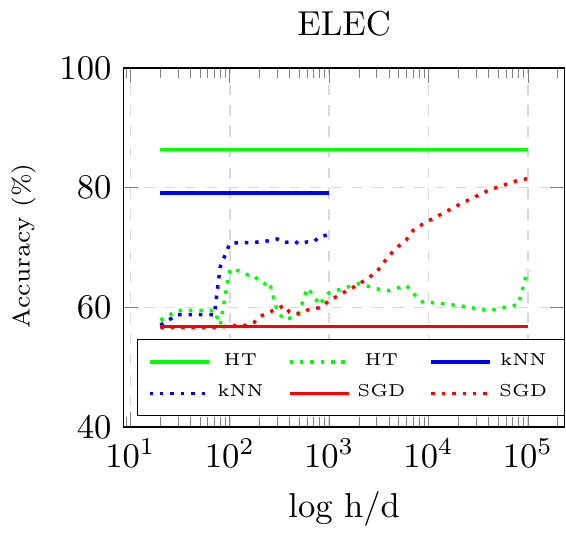}
	\includegraphics{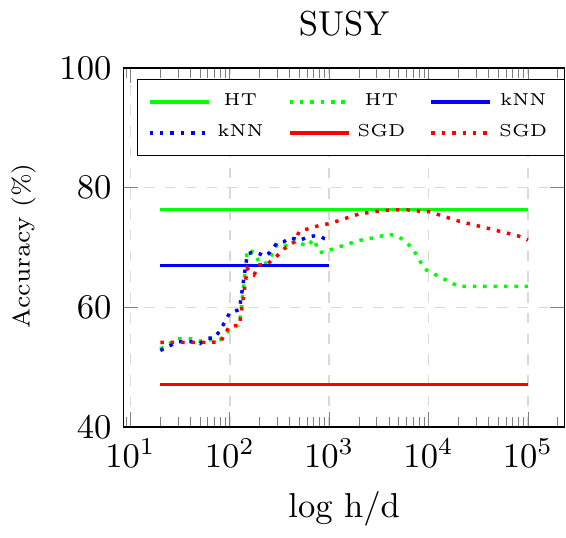}
    }\\
    \subfloat[test][Running Time (s)]{
	\label{fig:3b}
	\includegraphics{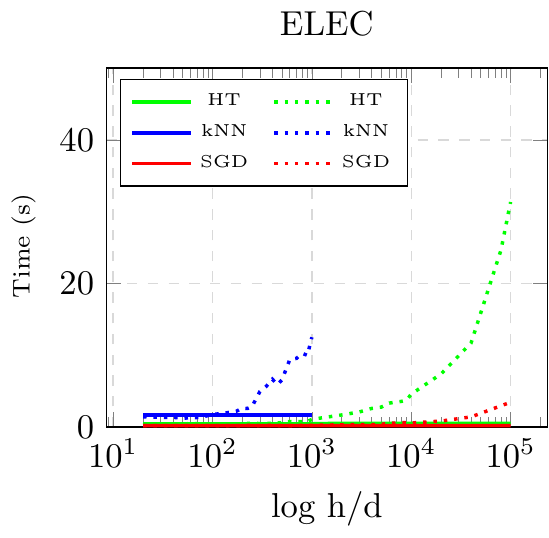}
	\includegraphics{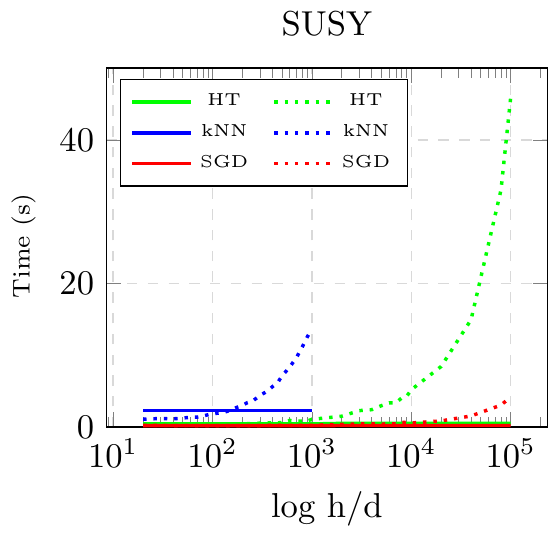}
    }
    \caption{\label{fig:3}Accuracy (\Fig{fig:3a}) and Running Time (\Fig{fig:3b}) on a 10,000-instance segment of two real-world datasets (\textsf{ELEC} and \textsf{SUSY}) for varying proportions of $h$ (number of hidden units / basis functions) wrt $d$. $k$NN has been cut out after $h=1000/d$ due to scalability reasons. Note the log scale on the horizontal axis.}
    \end{center}
\end{figure}

For the feature filter we used parameters $h=5d$ hidden units for $k$NN-F and $h=10d$ for SGD-F and HT-F (a decision based on the relative computational sensitivity of $k$NN to a larger attribute space -- for \textsf{LOC2} this means 25,000 attributes in the projected space for SGD-F, and half of that for $k$NN-F) -- except where this is varied in \Fig{fig:3}. For $k$NN we used a buffer size of 5000. For LB we specify 10 models. In other cases, the default parameters in MOA are used.

\begin{table}
	\centering
	\caption{\label{tab:datasets}Data sources used in the experimetnal evaluation. Synthetic datasets are listed first.}
\begin{tabular}{cccc}
	\hline
		Dataset     & \#Attributes& \#Instances   & \\
	\hline
		RBF1        &    10   &  100,000   &   \\ 
		HYP1        &   10      &  100,000   &   \\ 
		LED1        &    24     &  100,000   &   \\ 
		LOC1        &  25     &  100,000   &   \\
		LOC2        &  2500   &  100,000   &   \\
		Poker       &   10    &  829,201   &   \\
	\hline
		Electricity &    8    &  45,312    &    \\
		CoverType   &   54    &  581,012   &   \\
		SUSY        &   8     & 5,000,000  &   \\
	\hline
\end{tabular}
\end{table}

\Fig{fig:3} displays the results of varying the relative size of the new feature space (wrt to the original feature space) on two real-world datasets. Note that the feature space is different, so even when this ratio is $1:1$, performance may differ. 

With regard to $k$NN, performance improves with more feature functions. In one of the two cases, this is sufficient to overtake $k$NN on the original feature space. Unfortunately, $k$NN is particularly sensitive to the number of attributes, so complexity becomes an issue long before other methods. The new feature space does not help the performance of HT, and in neither case does it reach the performance of HT on the original feature space. In fact, it begins to decrease again. This is because too many features makes it difficult for HT to become confident enough to split on, and may split poorly. Also, by partitioning the feature space, interaction between the features is lost. 
SGD reacts best to a new feature space. As noticed earlier \cite{IDA2012}, SGD is a poor performer compared to HTs, however, working in a feature space of random ReLUs, SGD-F actually reaches HT performance (on \textsf{SUSY}, and looks promising under \textsf{ELEC}) with similar time complexity. Even at 1,000 times the original feature space, running time is acceptable (only a several seconds per 10,000 instances). On the other hand, the increased memory use is significant across all methods. SGD requires 1,000 times more memory in this setting.


\begin{table}
\begin{tiny}
  \newcommand{\alg}[1]{#1}		
  \caption{\label{tab:1}Final Accuracy and Running Times. The dataset-wise ranking is given in (parentheses) and the average of these ranks is given in the final row.}
  \begin{adjustwidth}{-0.5in}{-0.5in}
	\centering
	\subfloat[test][Accuracy]{
      \begin{tabular}{lrrrrrrrrr}
\hline
Dataset   & \alg{HT} & \alg{SGD} & \alg{kNN} & \alg{LB-HT} & \alg{LB-SGD-F} & \alg{kNN-F} & \alg{SGD-F} & \alg{HT-kNN} & \alg{HT-SGD-F} \\
\hline
RBF1       & 75.0 (5) & 54.5 (9) & 92.0 (2) & 88.7 (4) & 72.0 (8) & 90.4 (3) & 72.0 (7) & 92.6 (1) & 73.7 (6)  \\ 
RBFD       & 65.7 (5) & 51.3 (9) & 88.6 (1) & 79.5 (4) & 59.8 (8) & 86.3 (2) & 59.9 (7) & 84.9 (3) & 59.9 (6)  \\ 
HYP1       & 87.7 (1) & 50.3 (9) & 82.9 (4) & 85.7 (2) & 67.2 (7) & 77.0 (5) & 67.2 (7) & 83.3 (3) & 67.9 (6)  \\ 
LED1       & 73.1 (1) & 10.3 (9) & 62.8 (3) & 72.0 (2) & 15.6 (6) & 49.0 (5) & 15.5 (7) & 62.8 (3) & 15.5 (7)  \\ 
POKR       & 76.1 (6) & 68.9 (9) & 69.3 (8) & 87.6 (1) & 82.3 (2) & 81.5 (4) & 81.9 (3) & 74.8 (7) & 80.1 (5)  \\ 
LOC1       & 85.5 (8) & 80.4 (9) & 91.0 (2) & 90.5 (6) & 90.7 (4) & 88.8 (7) & 90.7 (3) & 91.3 (1) & 90.7 (5)  \\ 
LOC2       & 56.3 (5) & 51.5 (9) & 75.7 (2) & 52.6 (8) & 56.8 (4) & 74.5 (3) & 55.9 (7) & 75.9 (1) & 56.1 (6)  \\ 
\hline
ELEC       & 79.2 (4) & 57.6 (9) & 78.4 (5) & 89.8 (1) & 74.8 (7) & 74.2 (8) & 74.8 (6) & 82.5 (2) & 81.8 (3)  \\ 
COVT       & 80.3 (5) & 60.7 (9) & 92.2 (1) & 91.7 (2) & 78.7 (6) & 91.6 (3) & 78.7 (7) & 91.2 (4) & 78.3 (8)  \\ 
SUSY       & 78.2 (3) & 76.5 (7) & 67.5 (9) & 78.7 (1) & 77.7 (4) & 71.2 (8) & 77.7 (5) & 77.2 (6) & 78.4 (2)  \\ 
\hline
avg rank   &  4.30      &  8.80      &  3.70      &  3.10      &  5.60      &  4.80      &  5.90      &  3.10      &  5.40      \\
\hline
\end{tabular}

	}\\
	\subfloat[test][Running Time (s)]{
	\begin{tabular}{lrrrrrrrrr}
\hline
Dataset   & \alg{HT} & \alg{SGD} & \alg{kNN} & \alg{LB-HT} & \alg{LB-SGD-F} & \alg{kNN-F} & \alg{SGD-F} & \alg{HT-kNN} & \alg{HT-SGD-F} \\
\hline
RBF1       &   0 (3) &   0 (1) &   3 (6) &   3 (5) &   4 (8) &  14 (9) &   0 (2) &   4 (7) &   1 (4)  \\ 
RBFD       &   1 (3) &   0 (1) &   3 (6) &   2 (5) &   4 (8) &  15 (9) &   0 (2) &   4 (7) &   1 (4)  \\ 
HYP1       &   0 (2) &   0 (1) &   3 (6) &   2 (5) &   4 (7) &  13 (9) &   0 (3) &   4 (8) &   1 (4)  \\ 
LED1       &   0 (2) &   0 (1) &   7 (6) &   2 (5) &  17 (8) &  40 (9) &   1 (3) &   8 (7) &   1 (4)  \\ 
POKR       &   9 (2) &   3 (1) & 455 (8) &  91 (5) & 279 (6) & 1539 (9) &  21 (3) & 422 (7) &  26 (4)  \\ 
LOC1       &   1 (2) &   0 (1) &   8 (7) &   2 (5) &  21 (8) &  48 (9) &   1 (3) &   8 (6) &   2 (4)  \\ 
LOC2       &   9 (2) &   4 (1) & 1276 (7) &  93 (3) & 1917 (8) & 2270 (9) & 367 (5) & 1230 (6) & 350 (4)  \\ 
\hline
ELEC       &   1 (3) &   0 (1) &  14 (7) &  10 (6) &   9 (5) &  49 (9) &   1 (2) &  19 (8) &   2 (4)  \\ 
COVT       &  19 (2) &  11 (1) & 605 (6) & 220 (3) & 4119 (9) & 3998 (8) & 233 (4) & 727 (7) & 250 (5)  \\ 
SUSY       &  45 (2) &  25 (1) & 1464 (8) & 530 (5) & 1040 (6) & 4714 (9) & 118 (3) & 1428 (7) & 159 (4)  \\ 
\hline
avg rank   &  2.30      &  1.00      &  6.70      &  4.70      &  7.30      &  8.90      &  3.00      &  7.00      &  4.10      \\
\hline
\end{tabular}

	}
  \end{adjustwidth}
  
\end{tiny}

\end{table}

\begin{figure}
  \begin{center}
	\includegraphics{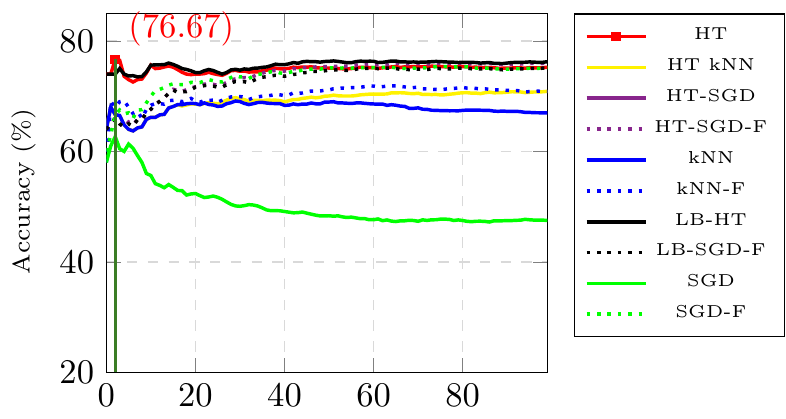}
	\includegraphics{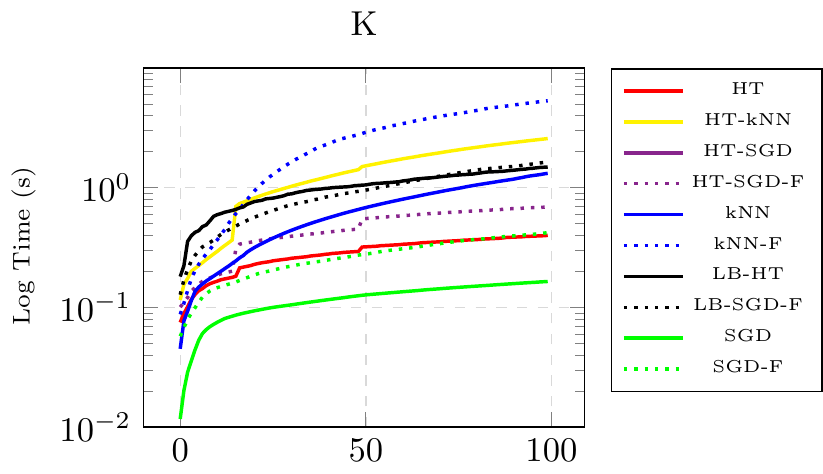}
    \caption{\label{fig:2}Performance over first 50,000 examples (right) of the \textsf{SUSY} data, in each divided into 100 windows.}
    \end{center}
\end{figure}

From this initial investigation we formulate several method combinations for a more extensive evaluation. \Tab{tab:1} displays the final accuracy over the data stream. The first four columns represent the baselines and state-of-the-art (LB-HT), and remaining columns are a selection of new method combinations. \Fig{fig:2} gives a more detailed over-time view of the largest dataset (\textsf{SUSY}), with the average performance plotted over the entire stream over 100 intervals, and also the first 1/10th of the data (again, over 100 intervals). The second plot gives more of an idea about how models respond to fresh concepts. Learning new concepts is a fundamental part in data streams of adapting to concept drift. 

Regarding this experiment some of the most important observations and conclusions are as follows:
\begin{itemize}
	\item SGD-F (i.e., SGD with random feature functions), even in this first analysis, out-competes established methods like $k$NN on several datasets. 
	\item $k$NN benefits relatively less (than SGD) from the feature functions filter. This is expected, since $k$NN is already a non-linear learner. 
		However, on a few datasets accuracy is 5-10 percentage points higher with the filter.
	\item $k$NN can be used effectively in the leaves of HT instead of the default of naive Bayes. There is an additional computational cost involved, but results showed this to be highly competitive method -- best equal overall in predictive performance tied with state-of-the-art LB-HT
	\item HT is difficult to improve on using feature functions (at least with the ReLUs that we experimented with). Again, this can be attributed to HT being a non-linear learner. Peak accuracy is reached in relatively short space of time.
	\item SGD takes longer than HT or LB-HT to reach competitive accuracy, but the gap narrows significantly with more examples (for example, under \textsf{SUSY}). On the largest datasets, the final average accuracy is within a percentage point -- and this average includes initial poorer performance. Therefore, on particularly big data streams (which are increasingly common), HTs could find themselves increasingly challenged to stay ahead of these methods.
	\item HT-SGD-F is comparable to the state of the art LB-HT on several datasets, but demonstrates more favourable running times.
	\item Unlike many deep learning techniques, these random functions do not require sensitive calibration.
	\item Unsurprisingly, $k$NN-based methods perform best on the dataset \textsf{RBFD} which has a drifting concept, since they automatically phase out older concepts. We did not look into detail about dealing with concept drift in this paper, but this can be dealt with by `meta methods', e.g., \cite{ADWIN}.
	\item Employing random feature functions as a `filter' in the MOA framework is a convenient and flexible way to apply it in a range of different data-stream methods.
\end{itemize}

%
%
\section{GPU Extended Evaluation} 
\label{sec:exp:dnn}

In the previous Section evaluations, we noticed SGD methods have the strongest advantage from random feature functions. This added to the increasing popularity of DL methods, we elected this strategy for further investigation and experimentation in this Section. A natural choice for implementing DNNs is to use GPUs to accelerate the calculations. Our experiments were evaluated on an NVIDIA Tesla K40c with 12GB of RAM each, 15 SMX and up to 2880 simultaneous threads and CUDA 7.0. 

Another motivation to use to GPUs the selection of the network hyper parameters by using cross-validation: for each dataset and activation functions different configurations are tested and best performing one is chosen. In turned out this was a high number of combinations and a way to accelerate the process is using GPUs.


The random projection layer is implemented using  an standard two layers feed-forward fully connected network. The input is fed to the random layer, which is never trained, and the output from this layer is forwarded to the trained layer. In this work we use the SGD and MSE as the training algorithm and objective function respectively for the last layer. 

We use three of the data sources from \Tab{tab:datasets} Covertype (COVT), Electricity (ELEC), and SUSY. This way we can compare the accuracy obtained in this Section against well known state-of-the-art algorithms.


\begin{table}
	\newcommand{\alg}[1]{#1}		
	\caption{\label{tab:gpu:layer:init} Random numbers initialization strategy for the different activation functions}
	\centering
	\begin{tabular}{l|c|c}
	\hline
    Activation &  Weight Matrix & Bias Vector \\
    \hline
	RBF & mean=0.0 and std=1.0 &  gamma \\
	Sigmoid & mean=0.0 and std=0.9 &   mean=0.0 and std=0.2 \\
	ReLU & mean=0.0 and std=1.0 &   mean=0.0 and std=0.1 \\
	ReLU Inc & mean=0.0 and std=1.0 &  0.0 \\
	\hline
	\end{tabular}
\end{table}


The initialization of each layer depends on the activation function used, we tried different random number initialization strategies and those for which we achieved the best results are summarized in  Table \ref{tab:gpu:layer:init}. Most of the weight matrices are initialized using random numbers with mean=0 and $\sigma=1.0$, except for the sigmoid activation function. The bias vector purpose and usage is activation function dependent.

Different activation functions have been tested at the random layer: RBF-gamma, Sigmoid, ReLU, incremental ReLU. Sigmoid and ReLU are used in the standard way. 
As we can see in Table \ref{tab:gpu:layer:init} bias vector for RBF stores the gammas, in out evaluations we use $\gamma = \{0.001, 0.01, 0.1, 1.0, 10.0\}$. ReLU incremental used the bias vector to store the incremental mean for each output attribute. At the trained layer always, we always use the standard sigmoid as the activation function. 

The same way as in Section \ref{sec:exp:tree}, the network is built incrementally using prequential learning; we visit each instance only one time. This is in contrast to typical DNNs training, where instances are loaded in batches and the algorithm iterates over them given number of times and, every time the error is reduced the model is checkpointed and spread to be used.

Table \ref{tab:gpu:res:best2} summarizes the best results we obtained, and it compares them with the best results obtained in Section \ref{sec:exp:tree} evaluations. We choose the algorithms by accuracy, and compared the time to run against them. Configuration were chosen by cross-validation using the following parameters: $\mu \in [0.1, 1.0]$ with an increment of $0.1$ and a similar range for learning rate. Sizes tested: $[10,100]$ increment of 10, $[100,1000]$ increment of 100, and two more sizes: 1500, 2000.

\begin{table}
\begin{scriptsize}
  \newcommand{\alg}[1]{#1}		
  \caption{\label{tab:gpu:res:best2}GPU tests; best results.}
  \begin{tabular}{l|ccc|cccc||c}
    \hline
     & \multicolumn{3}{c|}{Best Algorithms} & \multicolumn{4}{c||}{Random Projection} &\\
    \hline
	Dataset & Alg & Acc (\%) & Time (s) & Activ & Size  & Acc (\%) & Time (s) & speedup \\
	\hline
	ELEC    & LB-HT & \textbf{89.8} & 10 & Sigmoid & 100 & 85.33 & 1.2 & 8x \\
	COVT    & $k$NN & 92.2 & 605 & ReLU & 2000 & \textbf{94.59} & 32 & 17x \\
	SUSY    & LB-HT & \textbf{78.7} & 530 & RBF & 600 & 77.63 & 172 & 3x \\
	\hline 
  \end{tabular}	
\end{scriptsize}
\end{table}

For the electricity dataset random projection layer (RPL) obtained  an accuracy of 85.33\%  using a random layer of 100 neurons and a sigmoid activation function. As we can see in Table \ref{tab:gpu:res:best2} the best performing algorithm is the LB-HT which achieved a 89.8\%. If compared with results at Table \ref{tab:1}, we can see our method is the second best result, 4.47 percentage points less.

In the covertype dataset evaluation RPL obtained the best result for this dataset with an accuracy of 94.59\%, improving 2.39 percentage points the $k$NN algorithm using a ReLU activation functions.

Finally, our RPL performed relatively poorly in the SUSY dataset using 600 random neurons. We obtained a 77.63\% , 1.07 percentage points less than the LB-HT. This distance is lees than the distance obtained with the electricity algorithm, but if we rank out results with those in Table \ref{tab:1} we are at the sixth position.

With regard the time to complete, we can see the GPU is faster in all of the three datasets. For the electricity dataset RPL is 8 times faster, for the CoverType dataset 17 times faster and 3 times faster for the SUSY dataset.

Now we detail for each dataset the activation curves, the momentum and learning rate for this figure are the same across all sizes and  we used the ones for the best results to see how size affects the accuracy.

\subsection{Electricity Dataset}

\begin{table}
	\newcommand{\alg}[1]{#1}		
	\caption{\label{tab:gpu:res:elec}ELEC Evaluation}
	\centering
	\begin{tabular}{ccccr}
	\hline
    \alg{Activation} & \alg{Random Neurons} & \alg{$\mu$} & \alg{$\eta$} & \alg{Accuracy(\%)} \\
	\hline 
	SIG & 100 &  0.3 & 0.11 & \textbf{85.33} \\
	ReLU &  400 &  0.3 & 0.01 & 84.95 \\
	ReLU inc & 200 &  0.3 & 0.01 & 84.97 \\
	RBF $\gamma$=0.001 & 2000 &  0.7 & 1.01 & 72.13 \\
	RBF $\gamma$=0.01 & 2000 &  0.7 & 1.01 & 72.13 \\ 
	RBF $\gamma$=0.1 & 2000 &  0.7 & 1.01 & 72.13 \\
	RBF $\gamma$=1.0 & 2000 &  0.7 & 1.01 & 72.13 \\
	RBF $\gamma$=10.0 & 2000 &  0.7 & 1.01 & 72.13 \\
	\hline
	\end{tabular}
\end{table}

\begin{figure}
  \begin{center}
	\caption{Elec Dataset}
	\label{fig:gpu:elec:curves}
    \includegraphics{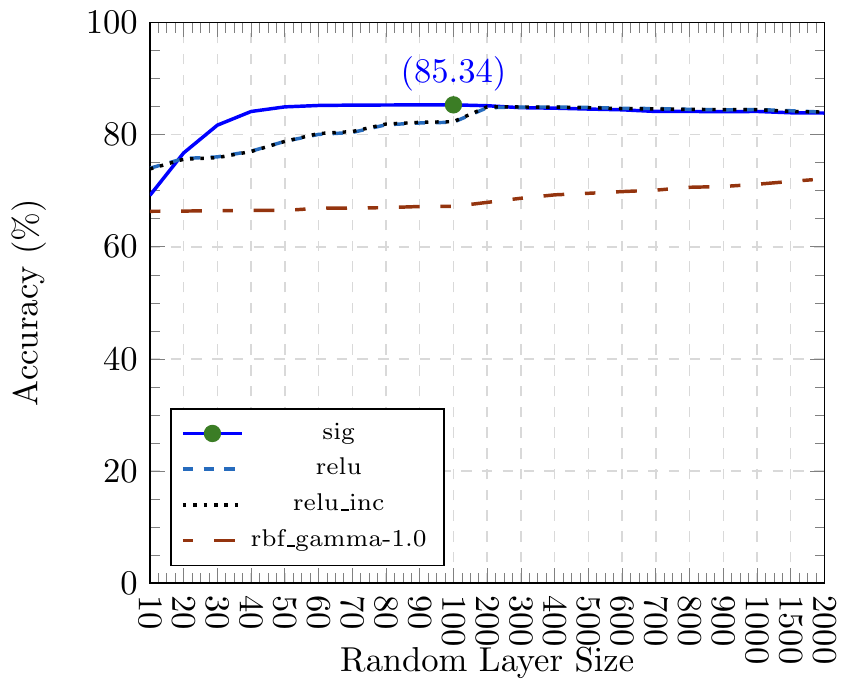}
     \end{center}
\end{figure}

Table \ref{tab:gpu:res:elec} summarizes the best results for each activation functions, and its configurations. As saw previously sigmoid activation function performed better than the others for this dataset. In the second position we find ReLU and ReLU inc activation functions which gave similar results, and slightly worse than the sigmoid. Regarding the RBF, all configurations we tried performed worse if compared to sigmoid, ReLU and ReLU inc, but very similar for the different gammas. In Figure-\ref{fig:gpu:elec:curves} we can see how accuracy changes with different sizes we tested. 

\subsection{CoverType Dataset}

\begin{table}
	\newcommand{\alg}[1]{#1}		
	\caption{\label{tab:gpu:res:cov:activs}COV Evaluation}
	\centering

	\begin{tabular}{rrrrrr}
	\hline
    \alg{Activation} & \alg{Random Neurons} & \alg{$\mu$} & \alg{$\eta$} & \alg{Accuracy(\%)} \\

	\hline 
	
	SIG & 1000 &  0.4 & 0.11 & 94.45 \\
	ReLU &  2000 &  0.4 & 0.01 & \textbf{94.59}\\
	ReLU inc & 2000 &  0.4 & 0.01 & 94.58 \\
    RBF $\gamma$=0.001 & 90 &  0.9 & 1.01 & 73.18\\
	RBF $\gamma$=0.01 & 90 &  0.9 & 1.01 & 73.18\\ 
	RBF $\gamma$=0.1 & 90 &  0.5 & 1.01 & 73.18\\ %
	RBF $\gamma$=1.0 & 90 &  0.8 & 1.01 & 73.18\\ 
	RBF $\gamma$=10.0 &  90 &  1.0 & 1.01 & 73.18\\ 
	\hline
	\end{tabular}
\end{table}

\begin{figure}
  \begin{center}

	\caption{COV Normalized Dataset}
	\label{fig:gpu:cov:curves}
    \includegraphics{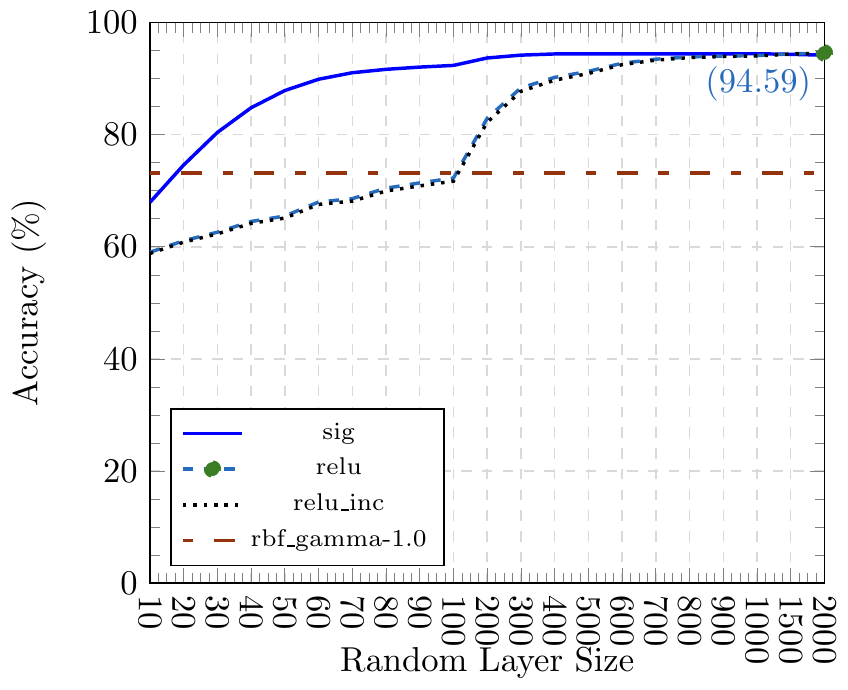}
    \end{center}
\end{figure}

Table \ref{tab:gpu:res:cov:activs} give us the best results for the COVT dataset each activation function. We can see a similar pattern as with the ELEC evaluation, SIG, ReLU and ReLU inc performed much better than the RBFs, and all three can beat results shown in Table \ref{tab:1}. This time the best result is obtained with the ReLU activation function at the random layer. 

In figure \ref{fig:gpu:cov:curves} we can see the activation curves. Although we got the best result with ReLU, the sigmoid has a better learning curve and it is very close to he ReLU accuracy. ReLU inc has a very similar learning curve as the standard ReLU. The different RBFs for the same momentum and learning with different sizes gives very similar (if not equal) results, so we chose the lower sizes.

\subsection{SUSY Dataset}

\begin{table}
	\newcommand{\alg}[1]{#1}		
	\centering
	\caption{SUSY Evaluation}
	\label{tab:gpu:res:susy:activs}

	\begin{tabular}{ccccr}
	\hline
   \alg{Activation} & \alg{Random Neurons} & \alg{$\mu$} & \alg{$\eta$} & \alg{Accuracy(\%)} \\
	\hline 
	SIG & 20 &   1 & 0.61 & 67.28 \\
	ReLU  &  20 & 1 & 0.61 & 74.84 \\
	ReLU inc &  20 & 1 & 0.91 & 74.80 \\ 
	RBF $\gamma$=0.001 & 600  & 1 & 0.71 & \textbf{77.63} \\
	RBF $\gamma$=0.01 &  600  & 1 & 0.71 & \textbf{77.63} \\
	RBF $\gamma$=0.1 & 600 & 1 & 0.71 & \textbf{77.63} \\
	RBF $\gamma$=1.0 &  600 & 1 & 0.71 & \textbf{77.63}\\
	RBF $\gamma$=10.0 &  600 & 1 & 0.71 & \textbf{77.63} \\
	\hline
	\end{tabular}
\end{table}

\begin{figure}
  \begin{center}
	\caption{SUSY Dataset}
	\label{fig:gpu:susy:curves}
	\includegraphics{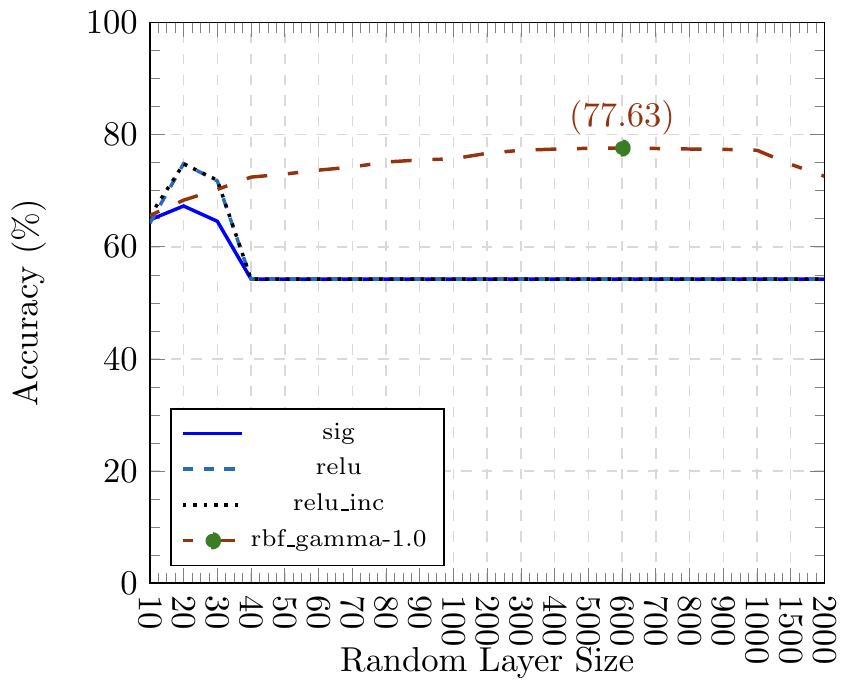}
    \end{center}
\end{figure}

Table \ref{tab:gpu:res:susy:activs} shows the best results for the SUSY dataset and activation function, and figure \ref{fig:gpu:susy:curves} the learning curves.
 The most noticeable effect is sigmoid, ReLU and ReLU inc the stop learning very soon, with only 20 random neurons ReLU reached its maximum peak with 74.85\%. The RBFs which performed poorly in previous evaluations, here are those with the best results. 

One curious result we can see is that the RBFs are performing around 7x\% in all 3 evaluations. Even if 2 of the 3 results are not very good, it seems they are not very sensitive to the different datasets, and somehow the results are stable across different data distributions. 

%
%
\section{Conclusions} 
\label{sec:conclusion}

In this paper, we studied combinations of Hoeffding trees, nearest neighbour, and gradient descent
methods adding a layer based on a random feature function filter.
We found that this random layer can turn a simple gradient descent learner into a competitive method for real-time data analysis. With this first attempt we could even improve on current state-of-the-art algorithms, scoring the best and the second best results for two out of three datasets tested. Like Hoeffding Trees and nearest neighbour methods, but unlike many many other gradient descent-based methods, the random layer works well without intensive parameter tuning.

We successfully extended and implemented on GPUs, obtaining powerful predictive performance. This suggests that using GPUs for data stream mining is a promising research topic for obtaining new fast and adaptive machine learning methodologies. 
 
In the future we intend to look for adding and pruning units incrementally in the stream over time to respond to make more efficient use of memory and adapt to drifting concepts. Also we would like to continue studying how to obtain new high scalable methods using GPUs.


\section*{Acknowledgment}

This work was supported in part by the Aalto University AEF research programme \url{http://energyefficiency.aalto.fi/en/}, by NVIDIA through the UPC/BSC GPU Center of Excellence, and the Spanish Ministry of Science and Technology through the TIN2012-34557. 

%
%

\bibliography{JournalOfSystemsAndSoftware}

\end{document}